\documentclass{article}

\usepackage{scalerel}
\usepackage{tikz}
\usetikzlibrary{svg.path}

\definecolor{orcidlogocol}{HTML}{A6CE39}
\tikzset{
  orcidlogo/.pic={
    \fill[orcidlogocol] svg{M256,128c0,70.7-57.3,128-128,128C57.3,256,0,198.7,0,128C0,57.3,57.3,0,128,0C198.7,0,256,57.3,256,128z};
    \fill[white] svg{M86.3,186.2H70.9V79.1h15.4v48.4V186.2z}
                 svg{M108.9,79.1h41.6c39.6,0,57,28.3,57,53.6c0,27.5-21.5,53.6-56.8,53.6h-41.8V79.1z M124.3,172.4h24.5c34.9,0,42.9-26.5,42.9-39.7c0-21.5-13.7-39.7-43.7-39.7h-23.7V172.4z}
                 svg{M88.7,56.8c0,5.5-4.5,10.1-10.1,10.1c-5.6,0-10.1-4.6-10.1-10.1c0-5.6,4.5-10.1,10.1-10.1C84.2,46.7,88.7,51.3,88.7,56.8z};
  }
}

\newcommand\orcidicon[1]{\href{https://orcid.org/#1}{\mbox{\scalerel*{
\begin{tikzpicture}[yscale=-1,transform shape]
\pic{orcidlogo};
\end{tikzpicture}
}{|}}}}

\usepackage[final]{neurips_2019}

\usepackage{url}
\usepackage{booktabs}
\usepackage{amsfonts}
\usepackage{nicefrac}
\usepackage{microtype,graphicx,subcaption,booktabs,soul,amsmath,bm,amsthm,xcolor,chapterbib,pifont}
\usepackage{hyperref}
\usepackage{pifont}
\hypersetup{urlbordercolor={white},}

\usepackage{cleveref}
\usepackage{xspace}
\usepackage[ruled,vlined]{algorithm2e}
\Crefname{algocf}{Algorithm}{Algorithms}

\DeclareMathOperator*{\diag}{\mathrm{diag}}

\newcommand{\kl}{\mathrm{KL}}
\newcommand{\js}{\mathrm{JS}}

\newcommand{\dx}{\mathrm{d}\bm{x}}

\makeatletter
\DeclareRobustCommand\onedot{\futurelet\@let@token\bmv@onedotaux}
\def\bmv@onedotaux{\ifx\@let@token.\else.\null\fi\xspace}
\def\eg{\emph{e.g}\onedot} 
\def\ie{\emph{i.e}\onedot}

\def\wrt{w.r.t\onedot}

\makeatother

\definecolor{orange}{RGB}{255,165,0}
\definecolor{lightcoral}{RGB}{240,128,128}
\definecolor{seagreen}{RGB}{46,139,87}
\definecolor{cornflowerblue}{RGB}{100,149,237}
\definecolor{silver}{RGB}{192,192,192}
\definecolor{darkviolet}{RGB}{148,0,211}
\definecolor{tomato}{RGB}{255,99,71}

\begin{document}

\title{Information-Geometric Set Embeddings (IGSE): From Sets to Probability Distributions
}

\author{Ke Sun~\orcidicon{0000-0001-6263-7355}\\CSIRO Data61\\Sydney, Australia\\sunk@ieee.org%
\And%
Frank Nielsen~\orcidicon{0000-0001-5728-0726}\\Sony Computer Science Laboratories, Inc.\\Tokyo, Japan\\Frank.Nielsen@acm.org}
\date{}
\maketitle

\begin{abstract}
This letter introduces an abstract learning problem called the ``set embedding'':
The objective is to map sets into probability distributions so as to lose less information.
We relate set union and intersection operations with corresponding interpolations of probability distributions.
We also demonstrate a preliminary solution with experimental results on toy set embedding examples.
\end{abstract}


%%%
\section{Set Embedding}
\begin{figure}[!h]
\includegraphics[width=\textwidth]{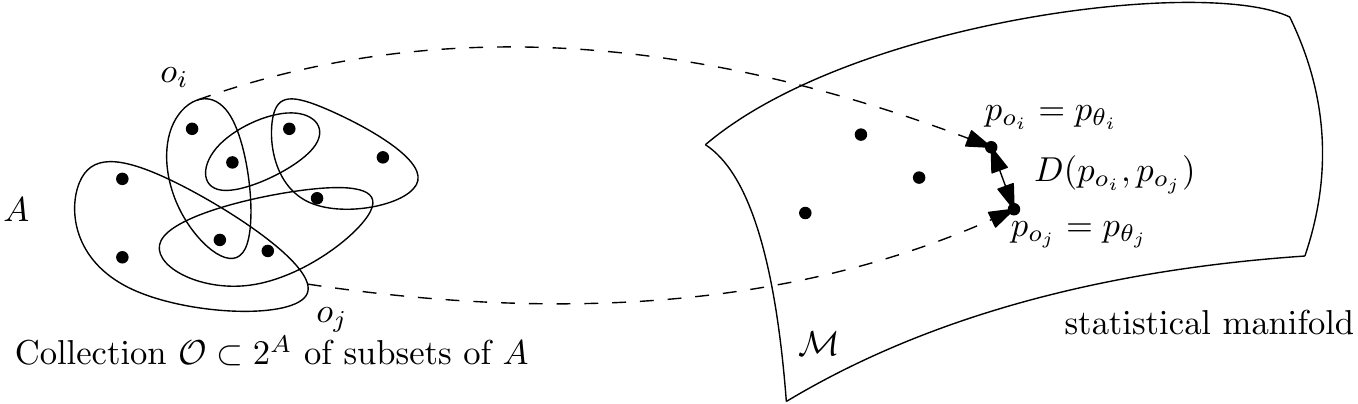}
\caption{Framework of information-geometric set embedding.}\label{fig:igse}
\end{figure}

%%%
The problem called {\em set embedding} is described as follows: We are given a collection
$\mathcal{O}\subset2^{A}$ of subsets (i.e., a \emph{family
of subsets}) of an {\em implicitly} given set $A$, where $2^A$ denotes the power set of $A$.
We aim to derive a faithful numerical representation of all elements of $\mathcal{O}$ by mapping
$\mathcal{O}$ to a continuous space $\mathcal{M}$, so that the images of subsets can
approximately preserve the relationships among the subset elements of $\mathcal{O}$.
See \cref{fig:igse} for an illustration of the general concept, where ``statistical manifold''
is a continuous space to be introduced latter.

This problem is interesting as a collection of subsets is a very basic algebraic structure,
and set embedding gives a uniform and continuous representation on a
non-uniform and possibly discrete inputs. Notice that we aim to embed the
elements of $\mathcal{O}$ that are subsets rather than elements of $A$ (thus we only need to know $A$ implicitly and can deal with infinite spaces).
One can construct singleton subsets to solve the later {\em element embedding} problem.
By definition, set embedding should be {\em permutation-invariant} because the function input
are subsets whose elements are not ordered.
In practice, this invariance can be implemented by special neural networks like DeepSets~\cite{deepsets}.

%\subsection*{Related Machine Learning Problems}

Set embedding is a broad problem and is closely related to {\em word and
sentence embeddings}~\cite{word2vec,sentense} (e.g., word2vec with vector space operations)
and {\em graph embedding}~\cite{paDOL,glNSF} (e.g., node2vec).
In the problems of word and sentence embeddings, a sentence can be regarded as an {\em ordered multiset}.
They can be extended to embedding partially ordered sets~\cite{partial}.
In graph embedding, a graph is a collection of edges, which is a collection of sets of two nodes.
In the case of hypergraphs~\cite{zjLWH}, an edge can be any non-empty set of vertices.
Set embedding is a more general concept than {\em hypergraph embedding},
because the vertex set in hypergraphs is usually finite, while in
sets the number of elements can be infinite.
The difference with previous work is also in terms of context:
set embedding can consider the ``metric'' properties of $A$ (\eg the volume of subsets)
that is often not considered by graph embedding.  It can even deal with
general \emph{topological spaces}, where the collection $\mathcal{O}$ of open subsets
is closed under certain operations.
The derived continuous representations can be useful in subsequent  downstream machine learning tasks.
Set embedding can also be useful for generating visualizations for researchers working on set theory
and theoretical computer science, and for making (intuitive) illustrations in textbooks.
If the given subsets are all finite, set embedding performs unsupervised multiple instance learning~\cite{mil}.

%%%%
\section{Information-Geometric Set Embedding}
%%%%
We further constrain the problem setting to {\em information-geometric
set embedding} (IGSE), where the embedding target space $\mathcal{M}$ is a statistical manifold~\cite{aIGI},
\ie a potentially curved space of probability distributions.
Hence, the set embedding problem reduces to find for each subset $X\in\mathcal{O}$ a corresponding probability distribution $p_X\in\mathcal{M}$.
Our considerations are listed as follows.

\begin{figure}[b]
\includegraphics[width=\textwidth]{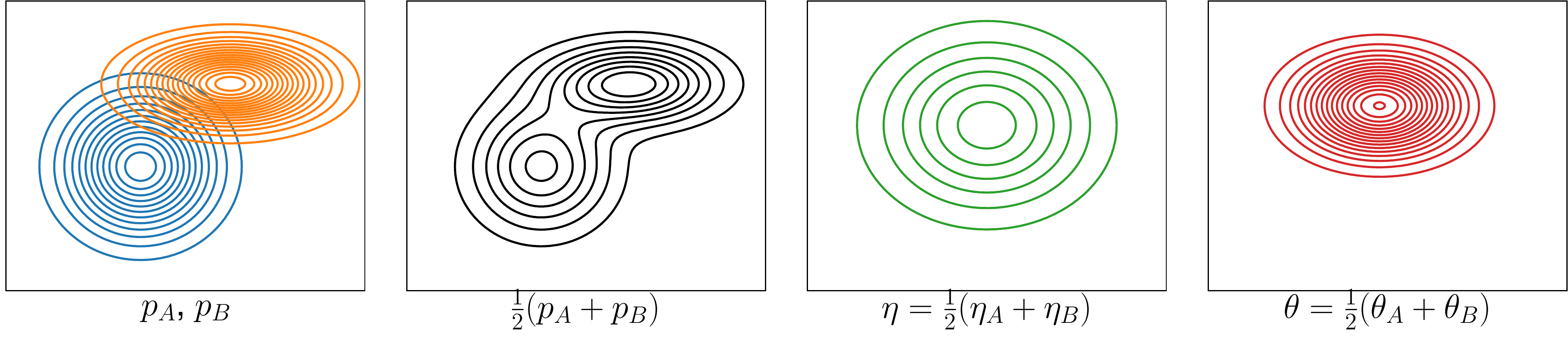}
\caption{From left to right: (1) Two Gaussian distributions; (2) their mixture;
(3) Their $m$-centroid (center of mass in the expectation parameters; mixture centroid tending to cover the support of the mixture);
(4) Their $e$-centroid  (center of mass in the natural parameters; exponential centroid tending to cover the mode of the mixture).}\label{fig:operation}
\end{figure}

\begin{itemize}

\item First, a statistical manifold
is a generalization of a flat Euclidean space. For example, consider the 2D space of
uni-variate Gaussian distributions with the coordinate frame
$(\mu,\sigma)$, where $\mu$ is the mean and $\sigma$ is the standard deviation.
The Fisher-Rao Riemannian geometry induced by the Fisher metric is of type hyperbolic like for any other location-scale family.
Notice that any subspace of constant standard deviation $\{(\mu,\sigma)\,:\,\sigma=\sigma_0\}$ is isometric to the real line~\cite{fimgauss}.
Therefore IGSE generalizes real vector embeddings.

\item Second, the cardinality $\vert{}X\vert$ of
a set $X$ naturally corresponds to the entropy $H(X)$ of a probability distribution $p_X$,
as they both measures the {\em uncertainty} in drawing a random element.
The most striking example is Hartley's entropy~\cite{hartley} defined by
$H_{\mathrm{Hartley}}(X)=\log\vert{X}\vert$, and thus $|X|=\exp\left(H_{\mathrm{Hartley}}(X)\right)$.

\item Third, informally, a distribution is a
``soft'' region of space characterized by its support, which bears some similarity to the concept of a set.

\item Fourth, basic set operations like union and intersection roughly corresponds to interpolating distributions.
A pair of distributions $p_A$ and $p_B$ in an exponential family can have
two different types of centroids: their {\em $m$-centroid} $c_m$, which is the linear
centroid in the expectation parameters $\eta$, or their {\em $e$-centroid} $c_e$, the
linear centroid in the natural parameters $\theta$ satisfying $c_\text{e}\propto\sqrt{p_Ap_B}$.
Another type of interpolation is taking the mixture model $p_\text{mix}=\frac{1}{2}(p_A+p_B)$,
which is generally outside of the exponential family (but stays inside for mixture families instead of exponential families).
These interpolating methods correspond to two basic set operations:
the union and the intersection. See {\em zero-forcing} (intersection) and {\em zero-avoiding} (union) properties
of left-sided/right-sided Kullback-Leibler centroids~\cite{symBD-2009} (see also~\cite{minka-2005} for properties with respect to $\alpha$-divergences).
We have $\mathrm{support}(p_\text{mix})\approx\mathrm{support}(A)\cup\mathrm{support}(B)$,
and $\mathrm{support}(c_e)\approx\mathrm{support}(A)\cap\mathrm{support}(B)$,
where $\mathrm{support}(\cdot)$ denote the {\em ``effective support,''} where the probability
density is sufficiently large.
See \cref{fig:operation} for a toy example of various interpolation schemes of two Gaussians.
\end{itemize}

We assume that the $\sigma$-algebra $(A,2^A)$ is associated with a probability measure $\mu$, so that
the uniform distribution can be defined.
We propose the following axioms that an IGSE method should (approximately) satisfy:

\begin{enumerate}
\item[\ding{192}] $\forall{X}\in\mathcal{O}$, the entropy $H(p_X)$ of $p_X\in\mathcal{M}$ is a monotonically increasing function of
$H(U_X)$, where $U_{X}$ means the uniform distribution over the elements of the subset $X\subset{A}$;

\item[\ding{193}] For $(X_1,X_2)$  a random pair of subsets in $\mathcal{O}^2$,
the statistic $D(p_{X_1} : p_{X_2})$ shall be positively correlated to
$D(U_{X_1} : U_{X_2})$, where $D$ is an {\em information divergence}~\cite{aIGI}
measuring dissimilarities between distributions.
\end{enumerate}

Informally, by the first condition, the capacity or uncertainty is preserved by IGSE.
By the second condition, the proximity of any pair of subsets is preserved.
That means, subsets with a large overlap are embedded close by, and subsets with
little or zero overlap are embedded far away.

Obviously, there exists a trivial set embedding satisfying both conditions:
the uniform distribution $U_X$. However, the uniform distribution is usually
not a compact representation and therefore is less useful: To describe the
uniform distribution, one needs to describe all elements in the subset $X$.
(This will require to consider the target domain to be $\Delta_{|A|}$,
the $|A|$-dimensional standard simplex parameterized by $\vert{A}\vert-1$ parameters.)
Instead, we constrain the target domain $\mathcal{M}$ to a parametric family of distributions with much less parameters,
\eg the space of Gaussian distributions, so as to derive a compact uniform
numerical representation (similar to the idea of dimensionality reduction).

If $\mathcal{O}$ has a finite cardinality, IGSE is reduced to the problem of embedding histograms.
We define the \emph{atomic subsets} \wrt $\mathcal{O}$ as $A_1,\cdots,A_m\in2^A$, so that
\begin{itemize}
\item $\forall{i}\in[m]$, $A_i\neq\emptyset$; if $i\neq{j}$, $A_i\cap{A_j}=\emptyset$; $\cup_{i=1}^mA_i = \cup_{X\in\mathcal{O}}X$;
\item $\forall{X}\in\mathcal{O}$, $\forall{i}\in[m]$,
$A_i\subset{X}$ or $A_i\cap{X}=\emptyset$.
\end{itemize}
The first condition means the atomic subsets form a \emph{partition} of
$\cup_{X\in\mathcal{O}}X$. By the second condition, if an atomic subset
has overlap with $X\in\mathcal{O}$, then it must be a subset of $X$.
For example, if $\mathcal{O}=\{X_1,X_2\}$,
$X_1-X_2\neq\emptyset$, $X_2-X_1\neq\emptyset$, and $X_1\cap{X}_2\neq\emptyset$, then
the atomic subsets are $X_1-X_2$, $X_2-X_1$, and $X_1\cap{X}_2$.
It is straightforward to prove that the set of atom subsets \wrt
a finite $\mathcal{O}$ is finite by mathematical induction.
One can recursively compute the atomic subsets of $\mathcal{O}$ as follows:
First, choose any $X\in\mathcal{O}$ and compute the atomic subsets $A_1,\cdots,A_{n}$
of $\mathcal{O}-X$. Then, the atomic subsets of $\mathcal{O}$ is given by all
non-empty subsets in $X-\cup_{i=1}^nA_{i}$, $A_1-X$, $\cdots$, $A_n-X$,
$A_1\cap{}X$, $\cdots$, $A_n\cap{}X$. In the worse case, the number of atomic
subsets grows exponentially \wrt $\vert\mathcal{O}\vert$.
The concept of atomic subsets yields an equivalence relation in $2^{2^A}$,
which contains all families of subsets:
If $\mathcal{O}_1$ and $\mathcal{O}_2$ induce the same atomic subsets,
then we denote $\mathcal{O}_1\sim_{\mathrm{atom}}\mathcal{O}_2$.
It means that the closures of $\mathcal{O}_1$ and $\mathcal{O}_2$ under
basic set operations (intersection, union, subtraction) are the same.
We propose the following invariance that an IGSE method should (try to) satisfy
\begin{enumerate}
\item[\ding{194}]
If $\mathcal{O}_1\sim_{\mathrm{atom}}\mathcal{O}_2$, then their IGSE should be
consistent, in the sense that in both embeddings $p_X$ is the same for any
$X\in\mathcal{O}_1\cap\mathcal{O}_2$.
\end{enumerate}

Then, $\forall{X}\in\mathcal{O}$, the associated $U_X$ can be defined as
a histogram over the atomic subsets. We iterate over the set of all
atomic subsets, and select the ones which satisfy $A_i\subset{X}$,
then $U_X$ is a mixture distribution
\begin{equation}\label{eq:decomp}
U_{X} = \frac{1}{Z} \sum_{i:A_i\subset{X}} V(A_i) U_{A_i},
\end{equation}
where $V(A_i)$ is the volume of $A_i$ \wrt the base measure $\mu$,
and $Z$ is the partition function. Therefore, the problem is
reduced to embedding histograms as other families of probability distributions.
This is different from information-geometric dimensionality reduction~\cite{crIGD,phd}
and distribution regression~\cite{distregression} in that both the source domain and
the target domain is a statistical manifold. Distribution
regression~\cite{distregression} is a supervised learning problem
and learns a mapping from a distribution, given implicitly by a set of random samples,
to a real valued response. Our problem setting is also different from information geometric kernel density estimation
(IGKDE; chapter 4~\cite{phd}) or embedding graph nodes into probability distributions~\cite{gauss}.

Notice that we can also associate to each element $a$ of $A$ an elementary Dirac probability distribution $p_a(x)=\delta_a(x)=\delta(x-a)$ ($1$ iff $x=a$ and $0$ otherwise), where the sample space of the distribution is $A$.
Then we view  a subset  $O\in\mathcal{O}$ as a mixture distribution of Diracs (``empirical distribution'' of the subset): $p_O(x)=\frac{1}{|O|} \sum_{a\in O} \delta_a(x)$, where $|\cdot|$ denote the cardinality of subset $O$.
Thus we reinterpret Eq.~\ref{eq:decomp} as a decomposition of a set mixture distributions into {\em atomic subset mixtures}.

Set embedding is different from graph drawing of Venn diagrams~\cite{VennDiagram-2003} that is a topological subset embedding.

%%%%
\section{A Preliminary Solution}
%%%%%
We describe a preliminary solution to solve IGSE through the Kullback-Leibler
(KL) divergence and the Jensen-Shannon (JS) divergence on the Gaussian manifold,
\ie the space of Gaussian distributions.
Our purpose is not a systematical empirical study, but to get some intuitions,
and to show that IGSE is easy to implement in practice.

By definition, the KL divergence is
$\kl(p : q) = \int p(\bm{x}) \left[\log{p}(\bm{x})-\log{q}(\bm{x})\right]\dx$.
It can be infinite if the input $U_{X_1}$ and $U_{X_2}$ have different supports.
We therefore use the \emph{damped KL} divergence
\begin{equation}
\kl_{\epsilon}(p : q) =
\max\left(
\int p(\bm{x}) \left[\log{p}(\bm{x})-\log\left(q(\bm{x})+\epsilon\right) \right]\dx,
\;0 \right),
\end{equation}
where $\epsilon>0$ is a small constant, to measure the distances between the
input uniform distributions.
For positive measures $\tilde{p}$ and $\tilde{q}$, we can define the extended Kullback-Leibler divergence as
\begin{equation}
\kl_{+}(\tilde{p}:\tilde{q}) = \kl(\tilde{p}:\tilde{q})+\tilde{q}-\tilde{p}.
\end{equation}
Thus $\kl_{\epsilon}(p : q)=\kl(p:q)+\epsilon \int \mu(x)\dx$, where the integral is calculated on the union of the support distributions.

For the output Gaussian distributions, their KL divergence is always well defined
if their covariance matrices have full rank.
Denote a pair of multivariate Gaussian distributions as
$G_1(\bm{x}\,;\,\bm\mu_1,\mathrm{diag}(\bm\sigma_1))$ and
$G_2(\bm{x}\,;\,\bm\mu_2,\mathrm{diag}(\bm\sigma_2))$,
where $\bm\mu_1$ and $\bm\mu_2$ are the means,
$\mathrm{diag}(\bm\sigma_1)$ and
$\mathrm{diag}(\bm\sigma_2)$ are the covariance matrices,
and
$\mathrm{diag}(\cdot)$ denotes the diagonal matrix
constructed using the given diagonal entries. We have
\begin{align}
&\kl(G_1:G_2) =
\sum_{j=1}^d \left[\log\sigma_2^{(j)}-\log\sigma_1^{(j)}
+ \frac{1}{2} \frac{(\sigma_1^{(j)})^2 + (\mu_1^{(j)}-\mu_2^{(j)})^2}{(\sigma_2^{(j)})^2}\right]
-\frac{d}{2},
\end{align}
where $\sigma_1^{(j)}$ denotes the $j$'th entry of the vector $\bm\sigma_1$,
and $d=\dim(\bm{x})$.
On the other hand, the Jensen-Shannon (JS) divergence~\cite{JS-2019} is
\begin{equation}\label{eq:js}
\js(p : q) =
\frac{1}{2}\kl\left( p(\bm{x}) : \frac{p(\bm{x})+q(\bm{x})}{2} \right)
+ \frac{1}{2}\kl\left( q(\bm{x}) : \frac{p(\bm{x})+q(\bm{x})}{2} \right),
\end{equation}
which is bounded in the range $[0,1]$ (when using base-$2$ logarithms) and can naturally handle the case when $p$
and $q$ have different supports.
To compute the JS divergence between two embedding points, we need to solve the
KL divergence between a Gaussian distribution and a Gaussian mixture of two
components. This KL divergence, on the RHS of \cref{eq:js}, by definition is an integration,
which can be approximated by Monte-Carlo sampling techniques and the reparameterisation
trick~\cite{vae}. Given $G_1$ and $G_2$, we have the approximation
\begin{align}
&\kl\left(G_1(\bm{x})\,:\,\frac{G_1(\bm{x})+G_2(\bm{x})}{2}\right)
\approx
-\sum_{j=1}^d \log\sigma_1^{(j)} -\frac{d}{2} + \log{2}\nonumber\\
&
-\frac{1}{K}\sum_{i=1}^K
\log\Bigg[
\exp\left(\sum_{j=1}^d\left( -\log\sigma_1^{(j)} -\frac{1}{2(\sigma_1^{(j)})^2}(x_i^{(j)}-\mu_1^{(j)})^2\right)\right)
\nonumber\\
&\hspace{5em}
+\exp\left(\sum_{j=1}^d\left( -\log\sigma_2^{(j)} -\frac{1}{2(\sigma_2^{(j)})^2}(x_i^{(j)}-\mu_2^{(j)})^2\right)\right)
\Bigg],
\end{align}
where $\{\bm{x}_i\}_{i=1}^K$ are i.i.d. samples drawn from $G_1(\bm{x})$.
At the limit $K\to\infty$, the approximation becomes accurate (i.e., consistent).
In summary, the divergence for all pairs in $\mathcal{O}$ and for all pairs
of embedding distributions can therefore be computed.

Given $\mathcal{O}$, we first augment it with the union of
\begin{align*}
\{X_1\cap{}X_2 \,:\, X_1, X_2\in\mathcal{O}^2\},\\
\{X_1\cup{}X_2 \,:\, X_1, X_2\in\mathcal{O}^2\},\\
\{X_1-X_2 \,:\, X_1, X_2\in\mathcal{O}^2\},\\
\{X_2-X_1 \,:\, X_1, X_2\in\mathcal{O}^2\},
\end{align*}
or a random subset of the union,
so that our IGSE has certain invariance \wrt set operations, as stated in
axiom \ding{194}. Note that it is hard to accurately satisfy \ding{194},
because there is an exponentially large number of
$\mathcal{O}'$ which satisfies $\mathcal{O}\sim_{\mathrm{atom}}\mathcal{O}'$.

Then, one can implement the IGSE through minimizing the {\em stress function}:
\begin{equation}\label{eq:stress}
\text{(axiom \ding{193})}\quad
\sum_{(X_1,X_2)\in\mathcal{O}^2} \Vert D(U_{X_1}:U_{X_2}) - a D(G_1: G_2)\Vert_2^2,
\end{equation}
with an auto-differentiation framework,
where $G_i$ is the Gaussian distribution associated with $X_i$,
and $a\in\Re^+$ is a free parameter.
In order to satisfying our axiom \ding{192},
we further constrain the covariance matrix of $G_{i}$ to be
$\diag(\bm\sigma_i)$, where
\begin{equation*}
\text{(axiom \ding{192})}\quad
\log \sigma_i^{(j)} = \tau^{(j)} + \log{}V_i,
\end{equation*}
where $\bm\tau\in\Re^d$ are free parameters, and $V_i$ is the volume or the
number of elements in $X_i$.  Intuitively, the larger the input subsets, the
larger the variance of the embedding distributions.
To avoid this reparameterisation of $\bm\sigma_i$,
an alternative method is to simply initialize the free parameter $\sigma_i^{(j)}$ with $V_i$.
%One can alternatively set $\tau$ to be a hyper-parameter for further constraints.
The computational complexity of the stress function is $O(Kd\vert\mathcal{O}\vert^2)$,
which can be further reduced by random sampling of the pairs
$(X_1,X_2)\in\mathcal{O}^2$.

\begin{table}[t]
\centering
\caption{Five different families of subsets}\label{tbl:set}
\begin{tabular}{r|c}
\hline
$\mathcal{O}_1$ & \colorbox{orange!50}{\{A\}}, \colorbox{lightcoral!50}{\{B\}}, \colorbox{seagreen!50}{\{C\}}, \colorbox{cornflowerblue!50}{\{A,B\}}, \colorbox{silver!50}{\{B,C\}}, \colorbox{darkviolet!50}{\{C, A\}}, \colorbox{tomato!50}{\{A,B,C\}}\\
$\mathcal{O}_2$ & \colorbox{orange!50}{\{A,B\}}, \colorbox{lightcoral!50}{\{B,C\}}, \colorbox{seagreen!50}{\{A\}}, \colorbox{cornflowerblue!50}{\{B\}}, \colorbox{silver!50}{\{C\}}\\
$\mathcal{O}_3$ & \colorbox{orange!50}{\{A,B,C,D,E,F\}}, \colorbox{lightcoral!50}{\{B,C,D\}}, \colorbox{seagreen!50}{\{C,D,E\}}, \colorbox{cornflowerblue!50}{\{A\}}\\
$\mathcal{O}_4$ & \colorbox{orange!50}{\{A,B,C,D,E,F,G\}}, \colorbox{lightcoral!50}{\{A,B,C,D,E\}}, \colorbox{seagreen!50}{\{A,B,C\}}, \colorbox{cornflowerblue!50}{\{A\}}\\
$\mathcal{O}_5$ & \colorbox{orange!50}{\{A,B,C\}}, \colorbox{lightcoral!50}{\{B,C,D\}}, \colorbox{seagreen!50}{\{C,D,E\}}, \colorbox{cornflowerblue!50}{\{D,E,A\}}, \colorbox{silver!50}{\{E,A,B\}}\\
\hline
\end{tabular}
\end{table}

\begin{figure}[t]
\centering
\begin{subfigure}[b]{.48\textwidth}
\includegraphics[width=\textwidth]{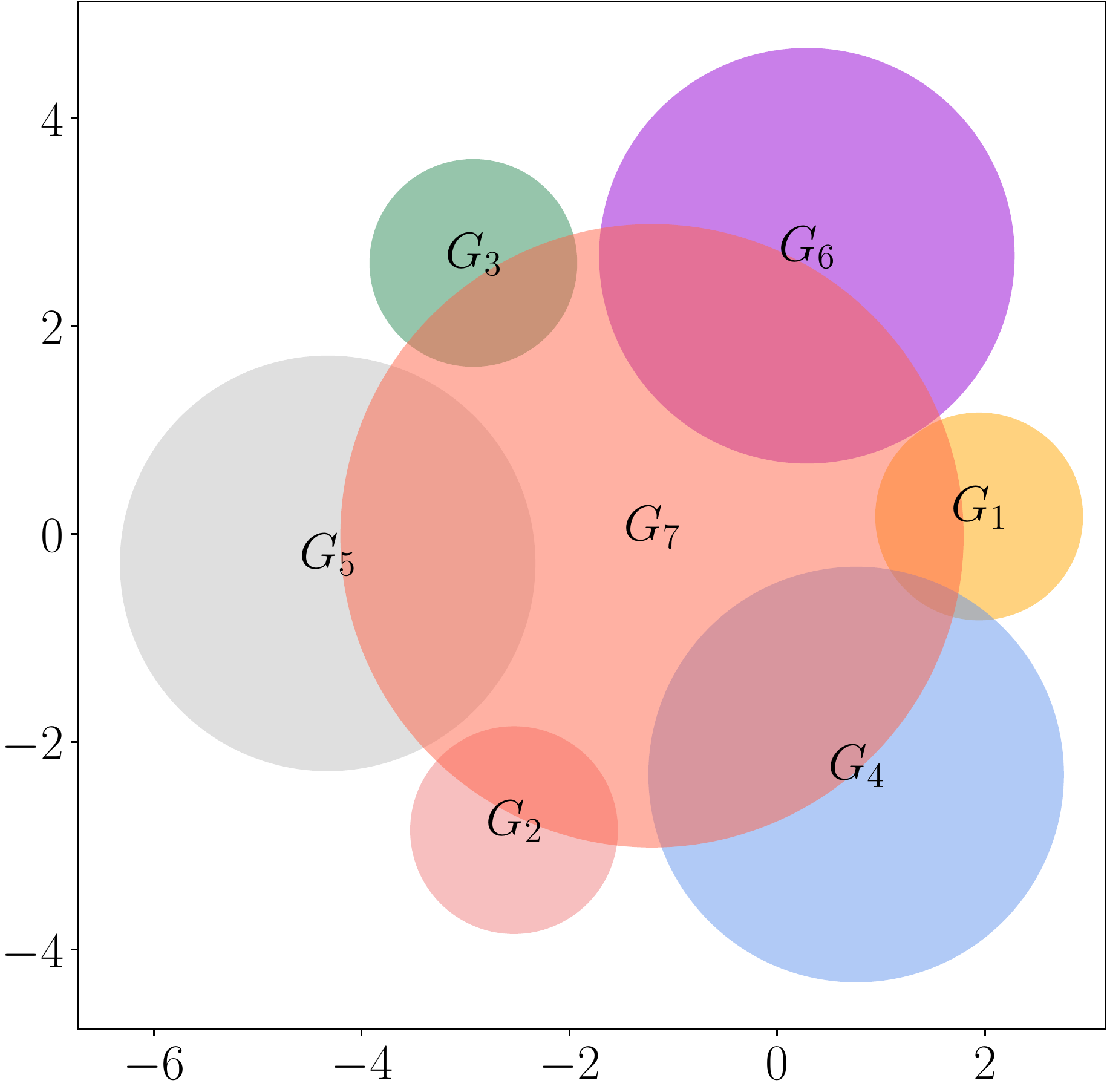}
\caption{KL ($\mathcal{O}_1$)}
\end{subfigure}
\begin{subfigure}[b]{.48\textwidth}
\includegraphics[width=\textwidth]{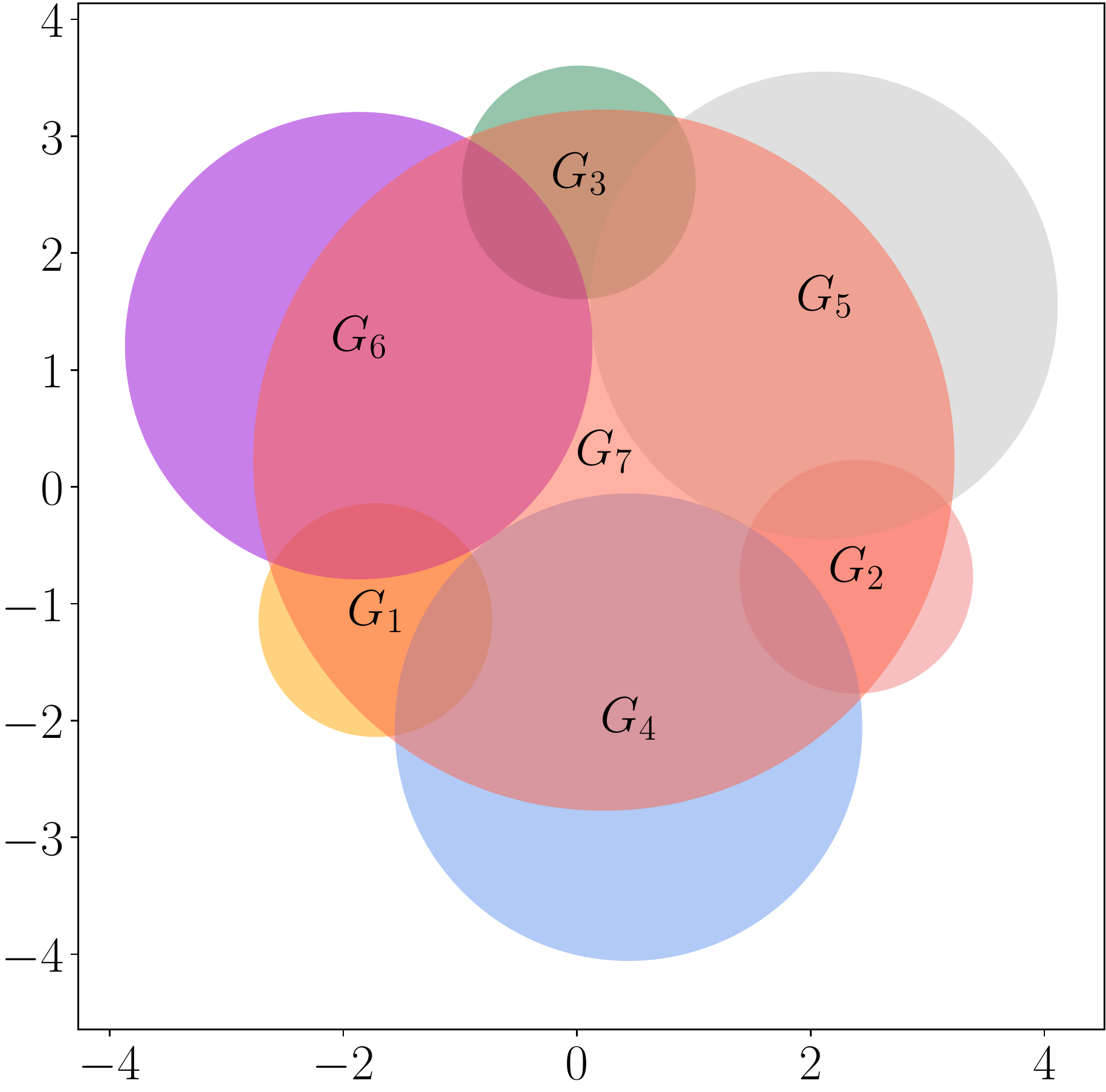}
\caption{JS ($\mathcal{O}_1$)}
\end{subfigure}

\begin{subfigure}[b]{.48\textwidth}
\includegraphics[width=\textwidth]{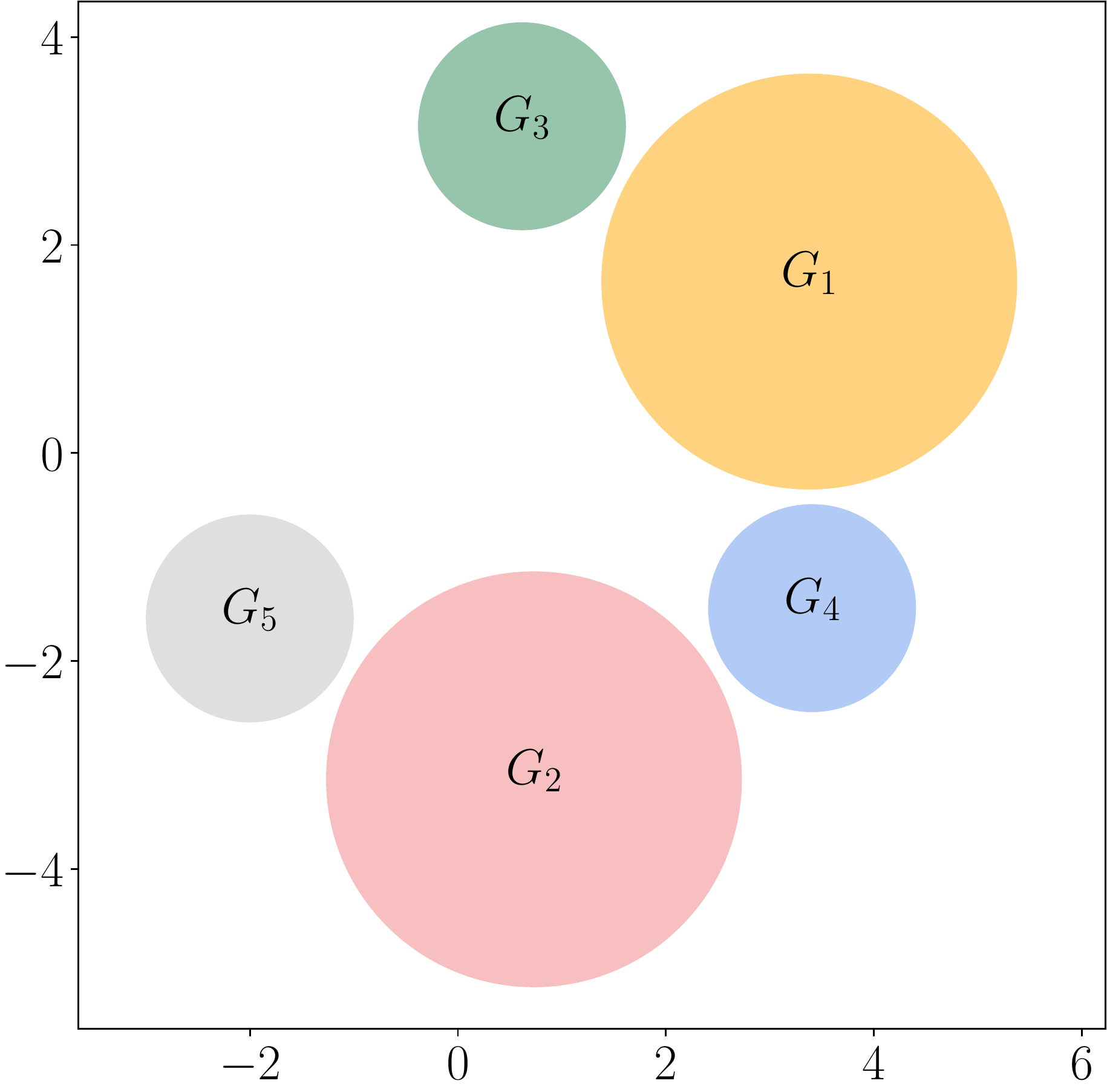}
\caption{KL ($\mathcal{O}_2$)}
\end{subfigure}
\begin{subfigure}[b]{.48\textwidth}
\includegraphics[width=\textwidth]{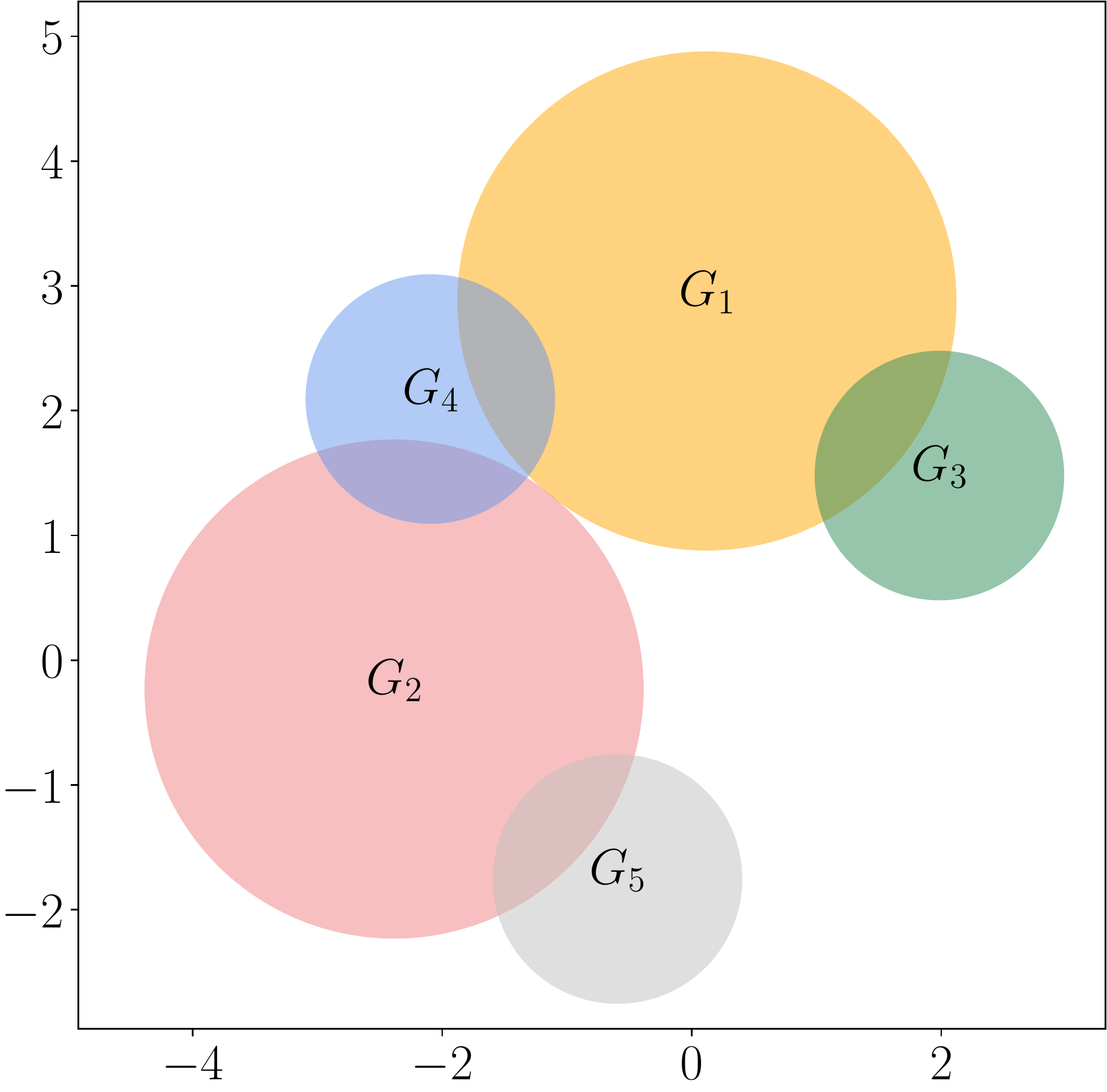}
\caption{JS ($\mathcal{O}_2$)}
\end{subfigure}

\begin{subfigure}[b]{.48\textwidth}
\includegraphics[width=\textwidth]{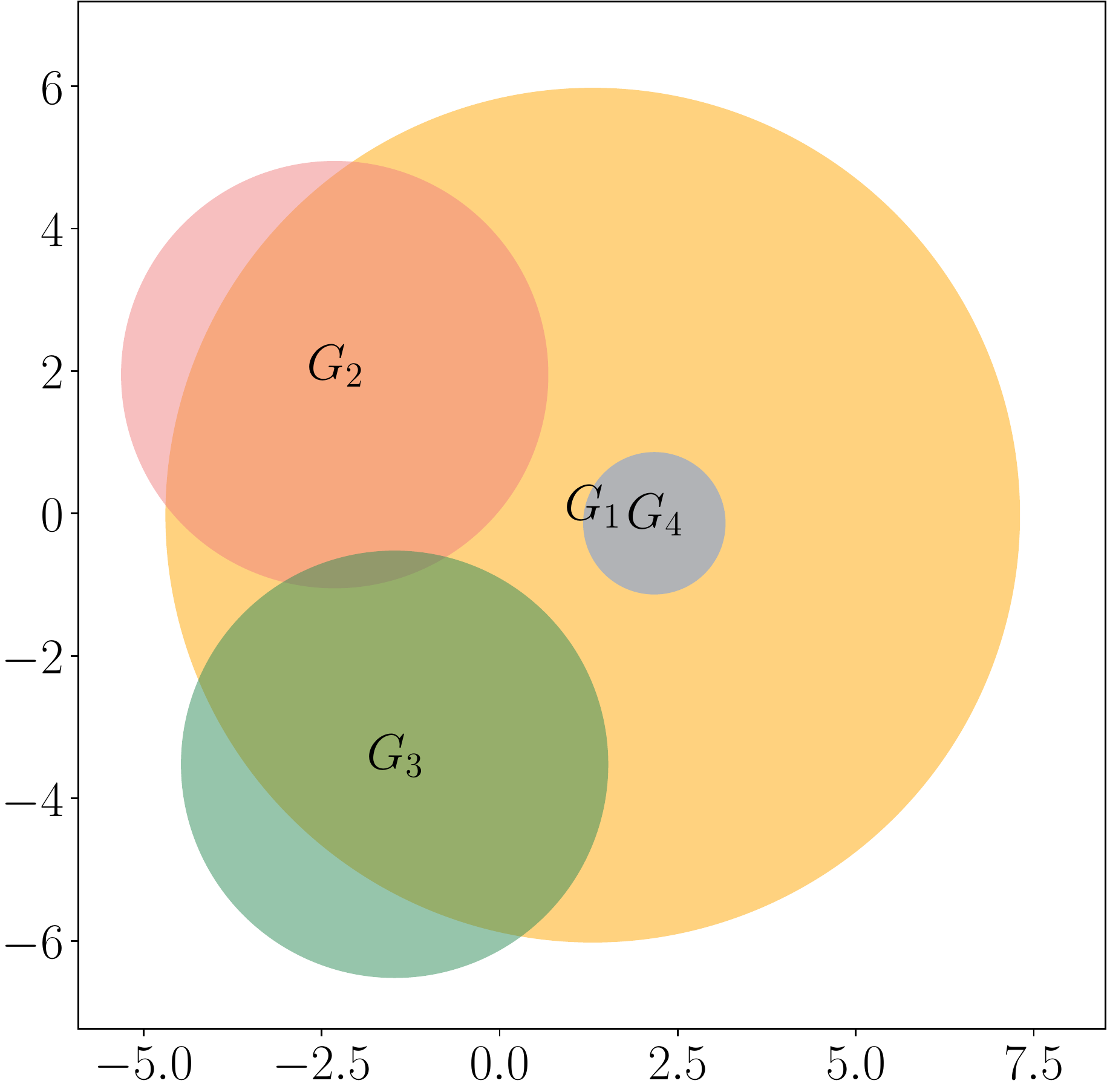}
\caption{KL ($\mathcal{O}_3$)}
\end{subfigure}
\begin{subfigure}[b]{.48\textwidth}
\includegraphics[width=\textwidth]{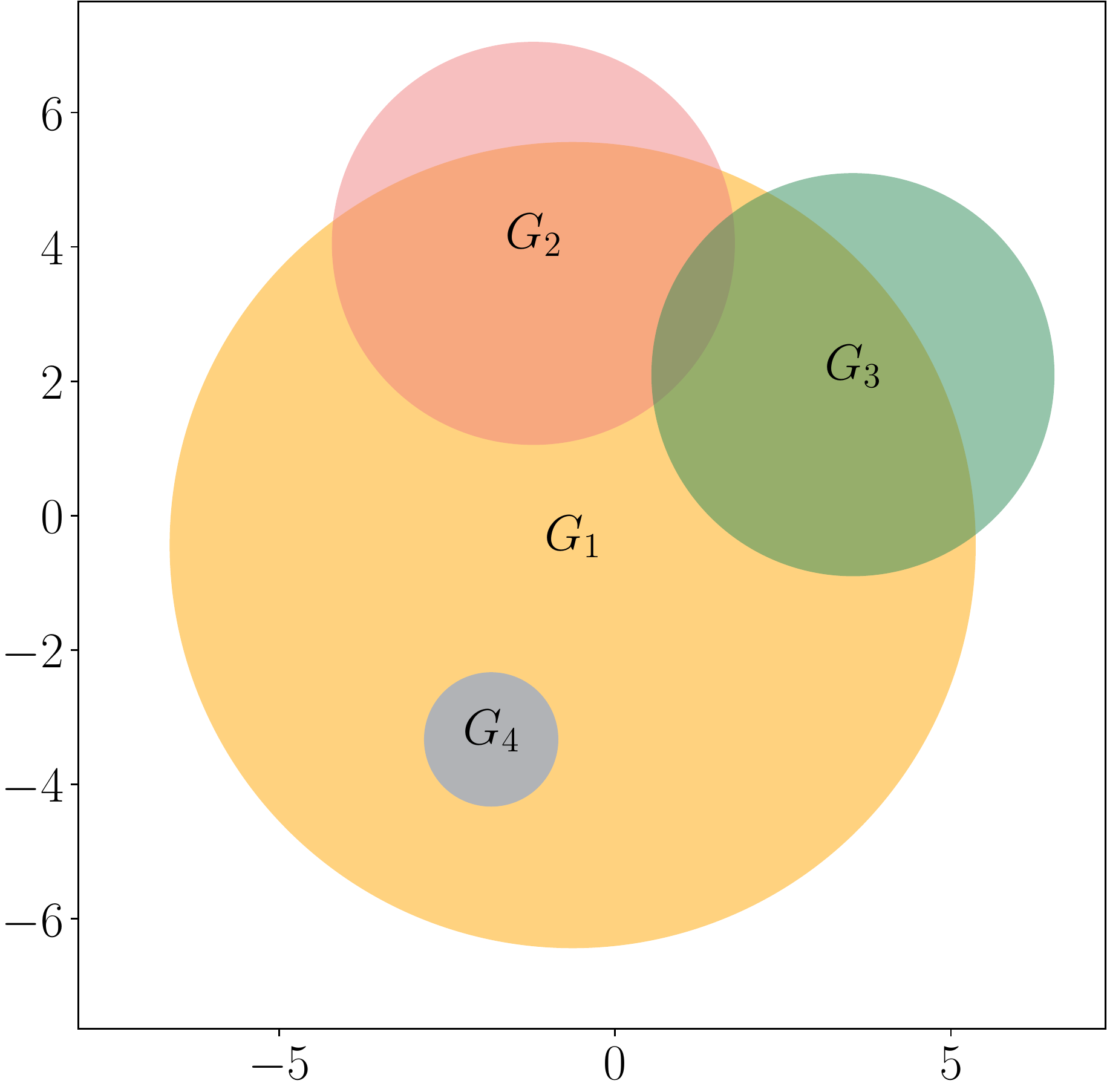}
\caption{JS ($\mathcal{O}_3$)}
\end{subfigure}

\caption{Embedding $\mathcal{O}_1$ (top), $\mathcal{O}_2$ (middle),
and $\mathcal{O}_3$ (bottom) into 2D Gaussian distributions based on
KL divergence (left) and JS divergence (right).
Each colored ellipse represents a 2D Gaussian distribution.
It shows the 2D region defined by mean$\pm$std.}\label{fig:toy}
\end{figure}

As a toy example,
we embed discrete subsets into 2D Gaussian distributions with diagonal
covariance matrices.
In order to minimize the stress function in \cref{eq:stress},
we apply batch gradient descent using the Adam optimizer~\cite{adam} based
on a constant learning rate. We use the set cardinality to initialize the
covariance matrix.
See \cref{tbl:set} for five families of subsets.
See \cref{fig:toy} for their corresponding 2D Gaussian embeddings.
For the first dataset $\mathcal{O}_1$, both embeddings based on the KL
and JS divergences can faithfully present the power set of $\{A,B,C\}$.
For $\mathcal{O}_2$, both embeddings reflect the relationships
between the given subsets. For example,
$G_4$ is in the middle of $G_1$ and $G_2$, showing the relationship
$\{\text{B}\}=\{\text{A},\text{B}\}\cap\{\text{B},\text{C}\}$.
For $\mathcal{O}_3$, both embeddings are similar and informative
\wrt the input subsets. For example, \{B,C,D\} (embedded into $G_2$)
and \{C,D,E\} (embedded into $G_3$) have a overlap.
For $\mathcal{O}_4$, JS appears better than KL, as it show a series of Gaussian
distributions with decreasing variance, and roughly contained in one another.
For $\mathcal{O}_5$, JS also appears better because it shows a circular
structure of the given subsets. Overall, our toy IGSE method based on two
different divergences can intuitively represent a given family of subsets, where
JS divergence seems to perform better as it can naturally handle distributions with
different support.

If the input contains not only a family $\mathcal{O}$ of subsets, but also
the features of the set elements, one should consider using deep neural networks
which are designed to be permutation invariant~\cite{deepsets,attention}. These
networks provide a \emph{parametric mapping} between the subsets and their
embedding images (distributions). This is different from the above
non-parametric approach, where the probability distributions are free parameters
to be learned. In this case, the neural network output should be a distribution
satisfying our axiom~\ding{192}, and the cost function should be designed to
satisfy our axiom~\ding{193}.
It is also possible to use such networks for the general case by feeding one-hot vectors as the input features.

\renewcommand{\thefigure}{\arabic{figure} (Cont.)}
\addtocounter{figure}{-1}
\begin{figure}[t]
\centering
\begin{subfigure}[b]{.48\textwidth}
\addtocounter{subfigure}{6}
\includegraphics[width=\textwidth]{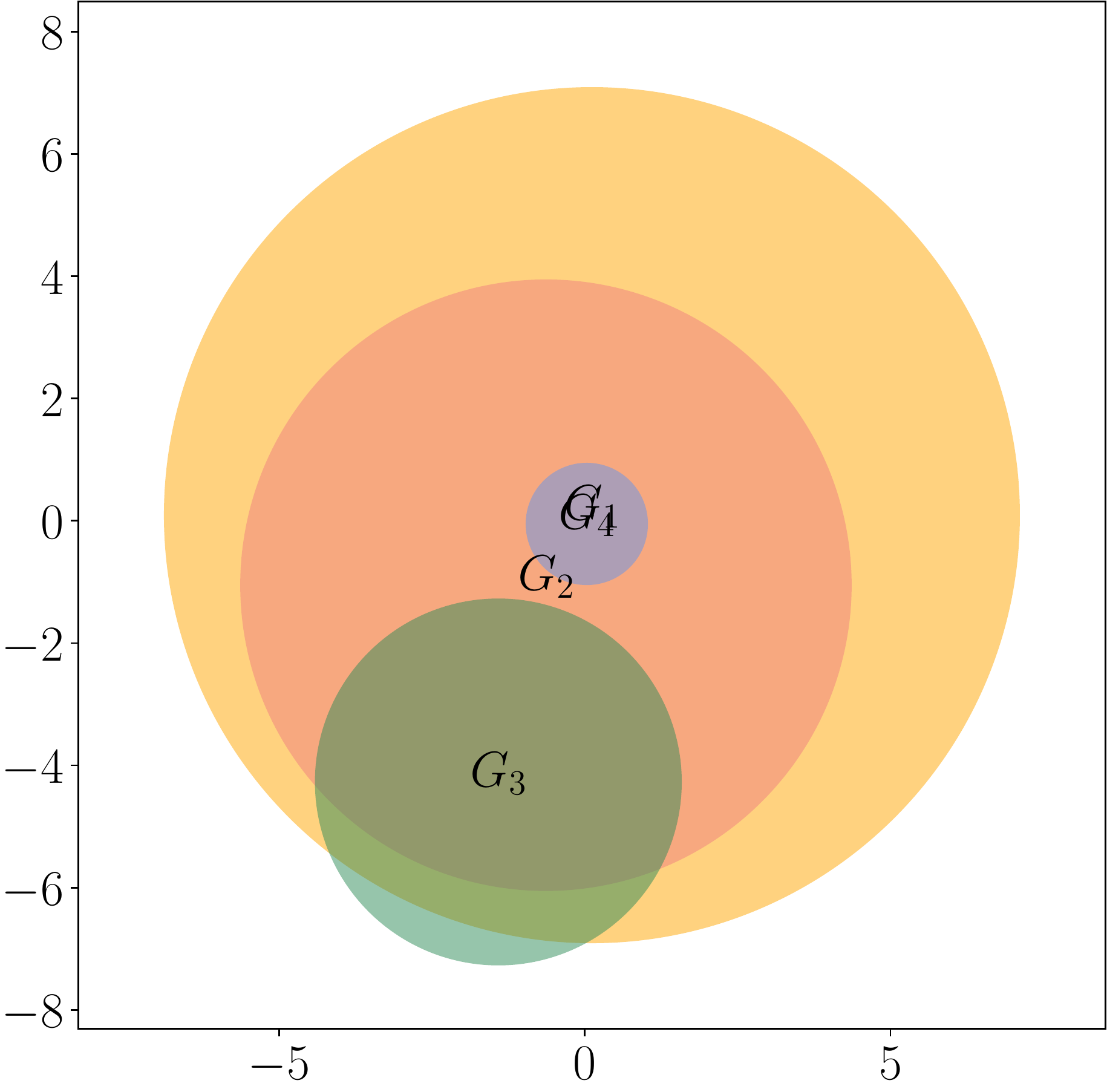}
\caption{KL ($\mathcal{O}_4$)}
\end{subfigure}
\begin{subfigure}[b]{.48\textwidth}
\includegraphics[width=\textwidth]{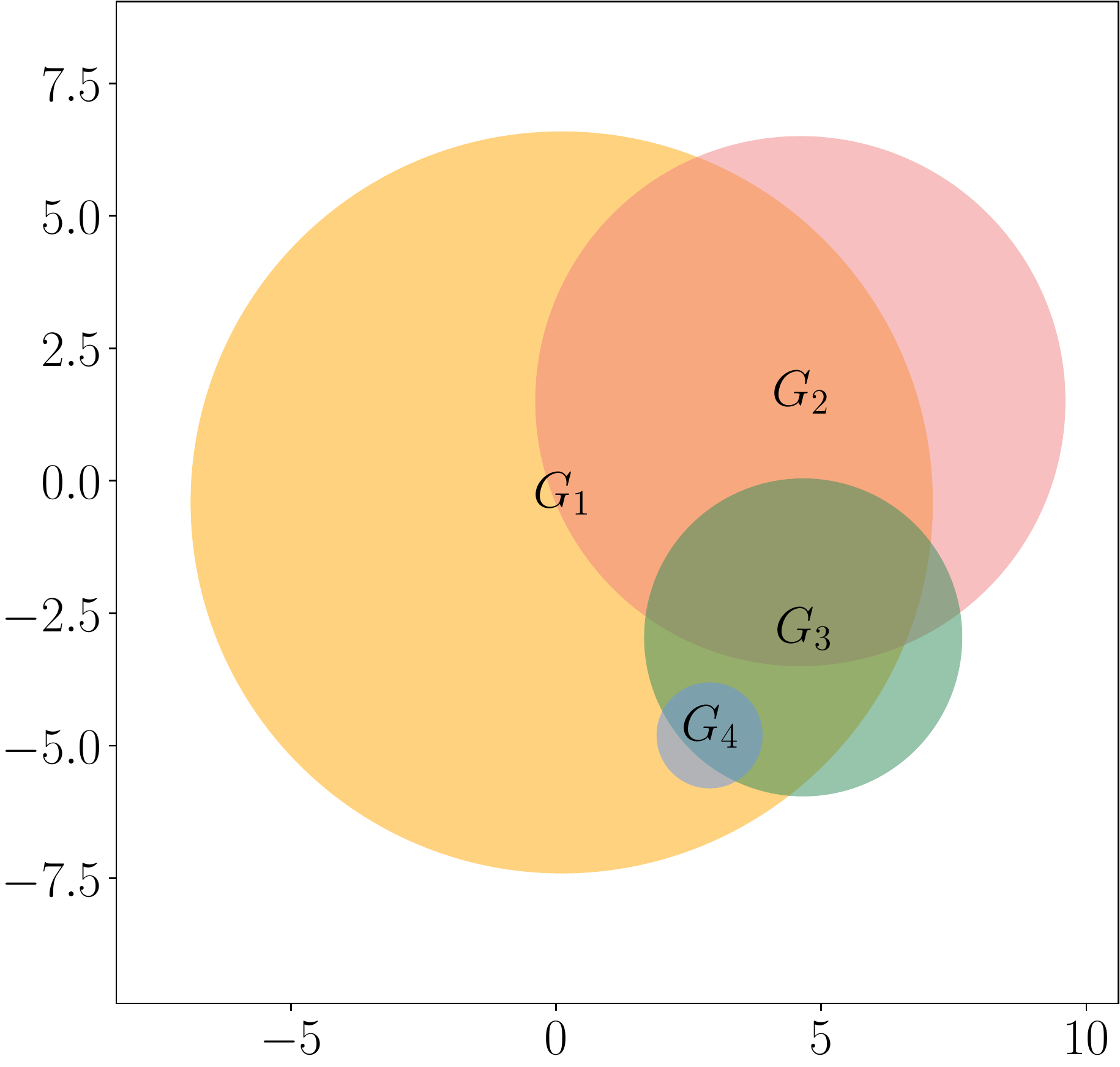}
\caption{JS ($\mathcal{O}_4$)}
\end{subfigure}
\begin{subfigure}[b]{.48\textwidth}
\includegraphics[width=\textwidth]{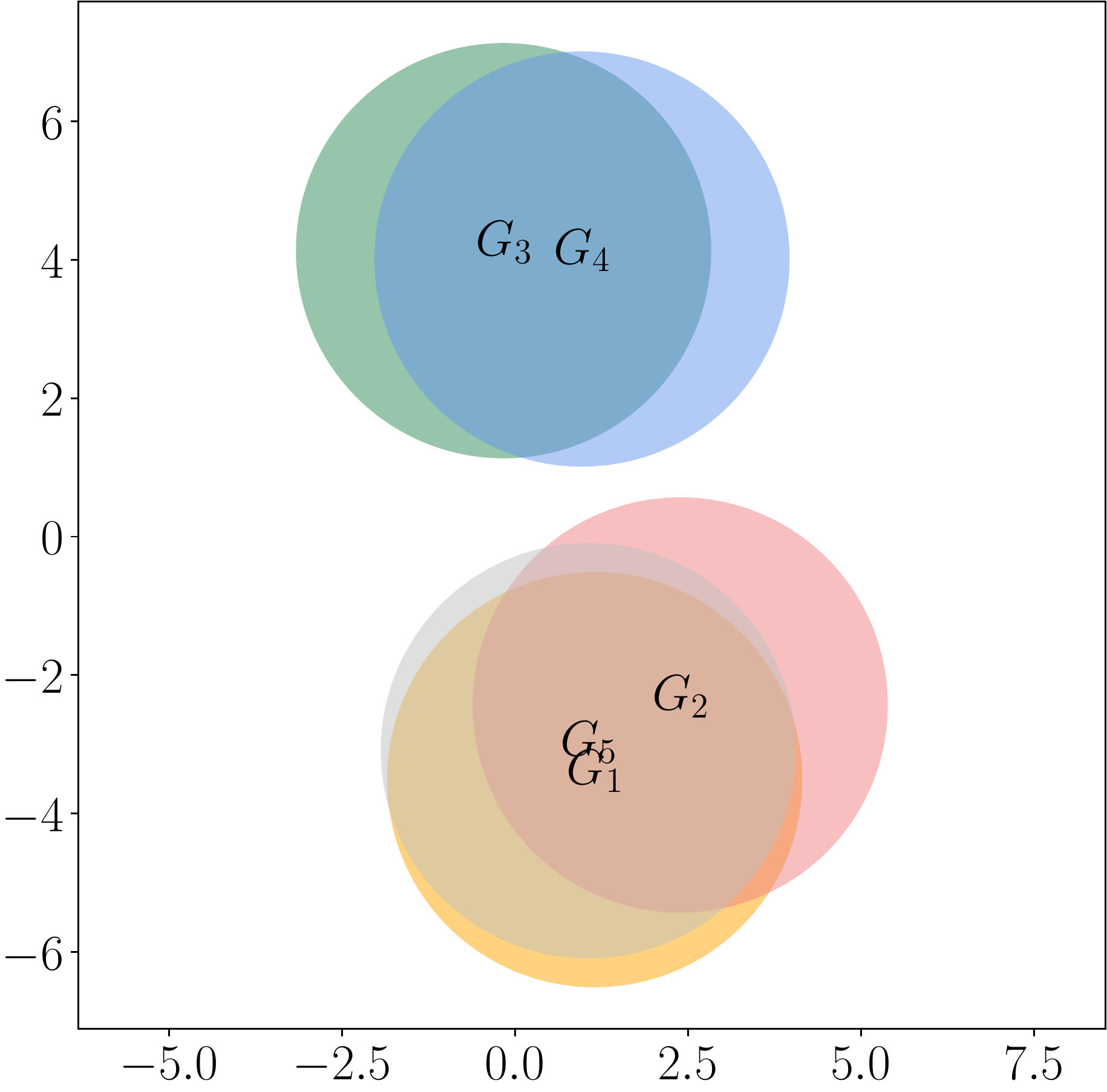}
\caption{KL ($\mathcal{O}_5$)}
\end{subfigure}
\begin{subfigure}[b]{.48\textwidth}
\includegraphics[width=\textwidth]{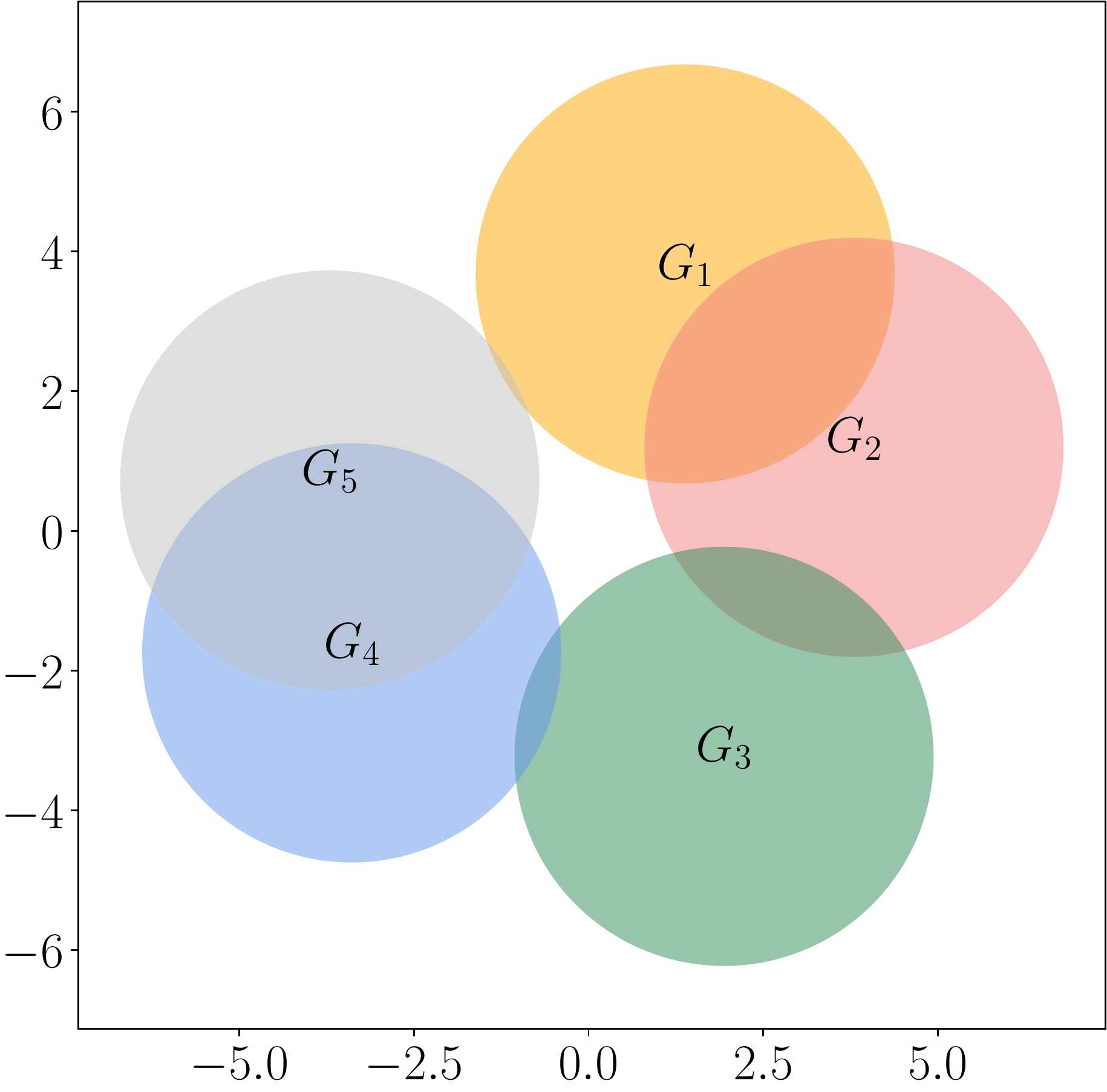}
\caption{JS ($\mathcal{O}_5$)}
\end{subfigure}
\caption{Embedding $\mathcal{O}_4$ and $\mathcal{O}_5$ into 2D Gaussian distributions based on
KL divergence (left) and JS divergence (right).}
\end{figure}
\renewcommand{\thefigure}{\arabic{figure}}

\subsubsection*{References}
\renewcommand{\refname}{}
\vspace{-2.5em}
\bibliographystyle{plain}
\bibliography{set}

\begin{thebibliography}{10}

\bibitem{aIGI}
S.~Amari.
\newblock {\em Information Geometry and Its Applications}, volume 194 of {\em
  Applied Mathematical Sciences}.
\newblock Springer-Verlag, Berlin, 2016.

\bibitem{partial}
M.~A. {Bautista}, A.~{Sanakoyeu}, and B.~{Ommer}.
\newblock Deep unsupervised similarity learning using partially ordered sets.
\newblock In {\em 2017 IEEE Conference on Computer Vision and Pattern
  Recognition (CVPR)}, pages 1923--1932, 2017.

\bibitem{gauss}
Aleksandar Bojchevski and Stephan Günnemann.
\newblock Deep {G}aussian embedding of graphs: Unsupervised inductive learning
  via ranking.
\newblock In {\em ICLR}, 2018.

\bibitem{mil}
Marc-Andr Carbonneau, Veronika Cheplygina, Eric Granger, and Ghyslain Gagnon.
\newblock Multiple instance learning.
\newblock {\em Pattern Recogn.}, 77(C):329--353, 2018.

\bibitem{crIGD}
K.~M. {Carter}, R.~{Raich}, W.~G. {Finn}, and A.~O. {Hero,III}.
\newblock Information-geometric dimensionality reduction.
\newblock {\em IEEE Signal Processing Magazine}, 28(2):89--99, 2011.

\bibitem{VennDiagram-2003}
Stirling Chow and Frank Ruskey.
\newblock Drawing area-proportional {V}enn and {E}uler diagrams.
\newblock In {\em International Symposium on Graph Drawing}, pages 466--477.
  Springer, 2003.

\bibitem{sentense}
Alexis Conneau, Douwe Kiela, Holger Schwenk, Lo{\"\i}c Barrault, and Antoine
  Bordes.
\newblock Supervised learning of universal sentence representations from
  natural language inference data.
\newblock In {\em Proceedings of the 2017 Conference on Empirical Methods in
  Natural Language Processing}, pages 670--680, 2017.

\bibitem{fimgauss}
Sueli I.~R. Costa, Sandra~A. Santos, and João~E. Strapasson.
\newblock Fisher information distance: a geometrical reading, 2012.
\newblock arXiv:1210.2354 [stat.ME].

\bibitem{glNSF}
A.~Grover and J.~Leskovec.
\newblock {Node2Vec}: Scalable feature learning for networks.
\newblock In {\em KDD}, pages 855--864, 2016.

\bibitem{hartley}
R.~V.~L. Hartley.
\newblock Transmission of information.
\newblock {\em Bell System Technical Journal}, 7(3):535--563, 1928.

\bibitem{attention}
Maximilian Ilse, Jakub Tomczak, and Max Welling.
\newblock Attention-based deep multiple instance learning.
\newblock In {\em Proceedings of the 35th International Conference on Machine
  Learning}, volume~80 of {\em Proceedings of Machine Learning Research}, pages
  2127--2136. PMLR, 2018.

\bibitem{adam}
Diederik~P. Kingma and Jimmy Ba.
\newblock Adam: A method for stochastic optimization.
\newblock In {\em ICLR}, 2015.

\bibitem{vae}
Diederik~P. Kingma and Max Welling.
\newblock Auto-encoding variational {B}ayes.
\newblock In {\em ICLR}, 2014.

\bibitem{word2vec}
Tomas Mikolov, Ilya Sutskever, Kai Chen, Greg~S Corrado, and Jeff Dean.
\newblock Distributed representations of words and phrases and their
  compositionality.
\newblock In {\em Advances in Neural Information Processing Systems 26}, pages
  3111--3119. Curran Associates, Inc., 2013.

\bibitem{minka-2005}
Tom Minka et~al.
\newblock Divergence measures and message passing.
\newblock Technical report, Technical report, Microsoft Research, 2005.

\bibitem{JS-2019}
Frank Nielsen.
\newblock On the {Jensen-Shannon} symmetrization of distances relying on
  abstract means.
\newblock {\em Entropy}, 21(5):485, 2019.

\bibitem{symBD-2009}
Frank Nielsen and Richard Nock.
\newblock Sided and symmetrized {B}regman centroids.
\newblock {\em IEEE transactions on Information Theory}, 55(6):2882--2904,
  2009.

\bibitem{paDOL}
B.~Perozzi, R.~Al-Rfou, and S.~Skiena.
\newblock {DeepWalk}: Online learning of social representations.
\newblock In {\em KDD}, pages 701--710, 2014.

\bibitem{phd}
Ke~Sun.
\newblock {\em Information Geometry and Data Manifold Representations}.
\newblock PhD thesis, Universit\'e de Gen\`eve, 2015.

\bibitem{distregression}
Zolt{{\'a}}n Szab{{\'o}}, Bharath~K. Sriperumbudur, Barnab{{\'a}}s
  P{{\'o}}czos, and Arthur Gretton.
\newblock Learning theory for distribution regression.
\newblock {\em Journal of Machine Learning Research}, 17(152):1--40, 2016.

\bibitem{deepsets}
Manzil Zaheer, Satwik Kottur, Siamak Ravanbakhsh, Barnabas Poczos, Ruslan~R
  Salakhutdinov, and Alexander~J Smola.
\newblock Deep sets.
\newblock In {\em Advances in Neural Information Processing Systems 30}, pages
  3391--3401. Curran Associates, Inc., 2017.

\bibitem{zjLWH}
Dengyong Zhou, Jiayuan Huang, and Bernhard Sch\"{o}lkopf.
\newblock Learning with hypergraphs: Clustering, classification, and embedding.
\newblock In {\em NIPS 19}, pages 1601--1608. MIT Press, 2007.

\end{thebibliography}

\end{document}